\documentclass[conference]{IEEEtran}
\IEEEoverridecommandlockouts

\usepackage{amsmath,amsthm,amssymb,amsfonts}

\usepackage{multirow}
\usepackage{colortbl}
\usepackage{float}
\usepackage[inline]{enumitem}

\usepackage{indentfirst}
\usepackage{array}
\usepackage[utf8]{inputenc}
\usepackage[most]{tcolorbox}
\tcbuselibrary{breakable}
\usepackage{listings}
\usepackage{xcolor}
\usepackage{longtable}
\usepackage{booktabs}
\usepackage{wasysym}
\usepackage{makecell}
\usepackage[linesnumbered, ruled]{algorithm2e}
\SetKwRepeat{Do}{do}{while}
\usepackage{booktabs}
\usepackage{alltt}
\usepackage{stfloats}
\usepackage{array}
\usepackage{mathtools}   
\usepackage{caption}
\usepackage{cite}
\usepackage{amsmath,amssymb,amsfonts}
\usepackage{algorithmic}
\usepackage{graphicx}
\usepackage{textcomp}
\usepackage{fontawesome}
\usepackage{xcolor}
\definecolor{gold}{RGB}{255,215,0}    
\definecolor{silver}{RGB}{192,192,192} 
\definecolor{bronze}{RGB}{205,127,50}  

\makeatletter
\DeclareRobustCommand\onedot{\futurelet\@let@token\@onedot}
\def\@onedot{\ifx\@let@token.\else.\null\fi\xspace}

\def\eg{\emph{e.g}\onedot} 
\def\ie{\emph{i.e}\onedot}

\makeatother

\usepackage{amsmath,amsfonts,bm}

\def\eqref#1{equation~\ref{#1}}

\def\1{\bm{1}}

\DeclareMathAlphabet{\mathsfit}{\encodingdefault}{\sfdefault}{m}{sl}
\SetMathAlphabet{\mathsfit}{bold}{\encodingdefault}{\sfdefault}{bx}{n}

\usepackage{hyperref}
\hypersetup{
    colorlinks=true,
    citecolor=blue,
    linkcolor=blue,
    filecolor=magenta,      
    urlcolor=cyan,
    pdfpagemode=FullScreen,
    }

\usepackage[capitalise]{cleveref}

\newcommand{\sysname}[0]{{\textsc{VeriRL}}\xspace}

\def\BibTeX{{\rm B\kern-.05em{\sc i\kern-.025em b}\kern-.08em
    T\kern-.1667em\lower.7ex\hbox{E}\kern-.125emX}}
\begin{document}

\title{\sysname: Boosting the LLM-based Verilog Code Generation via Reinforcement Learning
}

\author{
    Fu Teng$^{*1}$,
    Miao Pan$^{*1}$,
    Xuhong Zhang$^{\dagger1}$,
    Zhezhi He$^{\dagger2}$\thanks{$*$ indicates co-first authors. $\dagger$ indicates corresponding authors.},
    Yiyao Yang$^{2}$,\\
    Xinyi Chai$^{1}$,
    Mengnan Qi$^{2}$,
    Liqiang Lu$^{1}$,
    Jianwei Yin$^{1}$ \\
    $^1$Zhejiang University, China \quad
    $^2$Shanghai Jiao Tong University, China \\
    \{tengfu, zhangxuhong\}@zju.edu.cn
}

\maketitle


\maketitle

\begin{abstract}
Recent advancements in code generation have shown remarkable success across software domains, yet hardware description languages (HDLs) such as Verilog remain underexplored due to their concurrency semantics, syntactic rigidity, and simulation complexity. In this work, we address these challenges by introducing a reinforcement learning (RL) framework tailored for Verilog code generation. We first construct Veribench-53K, a high-quality dataset curated from over 700K Verilog problems, enriched with structured prompts, complexity labels, and diverse testbenches. To tackle the problem of sparse and noisy reward signals, we propose a Trace-back based Rescore mechanism that leverages reasoning paths and iterative refinement to enhance feedback reliability and support reward model training. Furthermore, to mitigate catastrophic forgetting and overfitting during RL fine-tuning, we introduce a sample-balanced weighting strategy that adaptively balances learning dynamics based on reward-probability distributions. These innovations are integrated into an iterative RL pipeline that co-evolves the policy and reward models. In contrast to recent work such as CraftRTL, which relies on large-scale closed-source model distillation, and DeepSeek-style approaches that struggle with sparse feedback, our method demonstrates superior performance using a smaller but high-quality dataset combined with RL optimization. Experiments on Verilog generation tasks demonstrate state-of-the-art performance, with substantial gains in test pass rate, functional correctness, and compilation robustness. Our findings highlight the potential of RL-driven approaches for structured code generation in hardware-centric domains.
\sysname is publicly available at \url{https://github.com/omniAI-Lab/VeriRL}.
\end{abstract}


\section{Introduction}

Large language models (LLMs) have advanced code generation, driven by scaling laws~\cite{kaplan2020scaling} and curated data~\cite{huang2024opencoder,guo2024deepseekcoderlargelanguagemodel}. State-of-the-art models like CodeLlama~\cite{rozière2024code}, Qwen2.5-Coder~\cite{hui2024qwen2}, and DeepSeek-Coder~\cite{guo2024deepseek} excel in program synthesis~\cite{chen2021evaluatinglargelanguagemodels}, test generation~\cite{steenhoek2023reinforcement}, and code repair~\cite{zheng2024opencodeinterpreterintegratingcodegeneration}. However, they still underperform on Hardware Description Languages (HDLs) like Verilog, due to challenges in concurrency, timing semantics, and strict syntax.

\begin{figure}[t]
    \centering
    \includegraphics[width=1\linewidth]{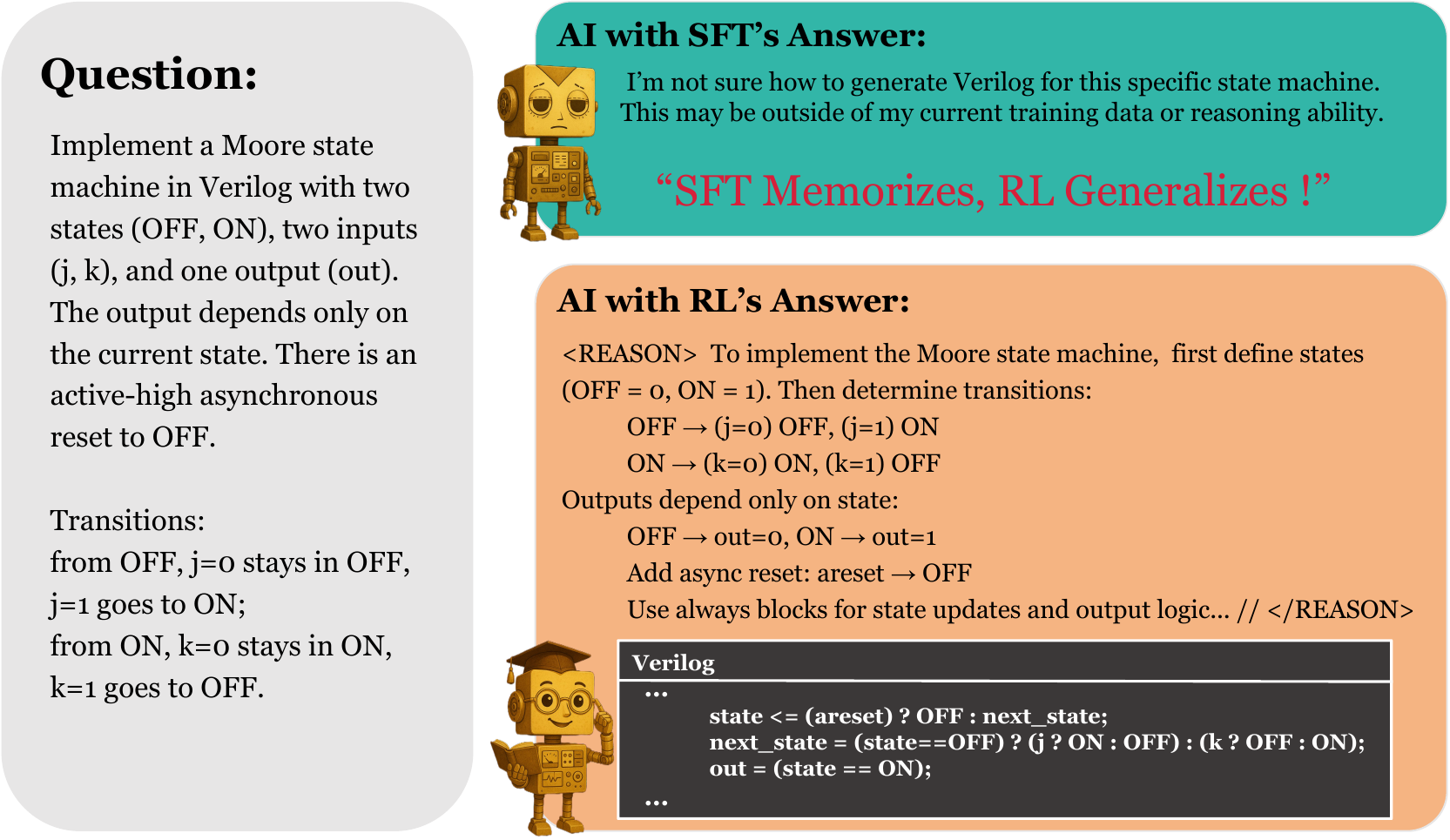}
    \caption{Supervised fine-tuning (SFT) versus RL based fine-tuning in Verilog code generation.}
    \label{highhigh}
\end{figure}

RTLFixer\cite{rtlfixer} and OriGen\cite{2024origen} have focused on enhancing
LLMs by reducing syntax errors. However, Verilog code generation demands more than syntactic fluency—it requires accurate reasoning over hardware semantics and robust simulation feedback. Moreover, previous works such as RTLCoder\cite{liu2024rtlcoder}, AutoVCoder\cite{autocoder2024} and OriGen\cite{2024origen}, have improved model performance by performing Supervised Fine-Tuning (SFT) using data synthesized by closed-source LLMs.
While SFT can learn surface-level patterns, it struggles to enforce executable correctness, particularly when high-quality data is scarce. In contrast, Reinforcement Learning (RL) offers an adaptive mechanism to optimize models toward functional goals by leveraging execution-based feedback\cite{gehring2024rlef}. More precisely, Fig \ref{highhigh} highlights the core distinction between SFT and RL fine-tuning: while SFT encourages pattern memorization, RL enables deeper reasoning by leveraging feedback from execution, thus guiding the model toward functionally correct solutions. This has shown promise in general-purpose languages through reward learning and test-guided rollouts \cite{chen2021evaluatinglargelanguagemodels, hu2025reinforce++}. However, applying RL to Verilog generation introduces two Challenges:
\begin{enumerate}[leftmargin=*]
\item \textit{Sparse and noisy reward signals}.\label{challenge1} Verilog lacks standardized simulation infrastructure. Building testbenches is complex due to concurrency and timing, and poor initial model performance leads to unstable, often binary feedback signals, hindering stable policy learning.

\item \textit{Catastrophic forgetting and reward variance}.\label{challenge2} Limited data and rigid syntax make RL fine-tuning prone to overfitting recent rewards and forgetting core logic patterns. Existing stabilization techniques (e.g., KL penalties, replay buffers) from RLHF are less effective for Verilog without task-specific reward design and update control.
\end{enumerate}

To tackle these issues, we propose a comprehensive framework that integrates high-quality data, structured reward modeling, and robust RL training, tailored for Verilog. We first construct \emph{Veribench-53K}, a curated dataset extracted from the large-scale PyraNet corpus of 700K Verilog problems \cite{nadimi2024pyranet}. We apply MinHash-based deduplication, compilation filtering, and quality screening (PyraNet's original score and complexity labels), preserving only Intermediate to Expert problems with a score $\geq$ 12. Each sample is further augmented through: (i) structured VerilogEval-human-style prompt \cite{liu2023verilogeval} reformulation to clarify design intent, and (ii) testbench synthesis, generating 5-10 diverse testbenches per problem covering edge sensitivity, timing, and functionality.

Next, we proposed a Trace-back based Rescore mechanism (TbR) to solve Challenge \ref{challenge1}. The wrong samples are not discarded directly, but selectively expanded according to the testbench pass rate. Then the model generates multiple reasoning paths for the selected wrong samples, which will get rescored by the new pass rate. Finally, the rewards are normalized to reflect the complexity of reasoning. This structured exploration greatly enriches the reward signal and provides reliable supervision for training reward models. To solve Challenge \ref{challenge2}, we propose a Sample-balanced Weighting (SbW) strategy, categorizing rollouts by their reward-probability dynamics and adjusting their gradient contributions. This reduces variance and mitigates forgetting.

While prior work~\cite{liu2024craftrtl} achieves strong results via SFT on large-scale distilled data from closed-source LLMs (e.g., ChatGPT), we take a different approach: leveraging a smaller dataset with a tailored RL framework. By integrating Veribench-53K, a trace-back reward mechanism, and a sample-balanced weighting strategy for stable fine-tuning, our method consistently outperforms SFT-only pipelines across diverse Verilog synthesis tasks. Our main contributions are:
\begin{itemize}[leftmargin=*,label={$\triangleright$}]
\item In this work, we introduce \sysname which applies advanced reinforcement learning based training upon LLM for Verilog code generation. To our best knowledge, \sysname is first work of such kind, which achieves state-of-the-art peferformance over prior work. 

\item We introduce a high-quality dataset with structured Verilog prompts and diverse testbenches, enabling scalable training and evaluation, as introduced in \cref{sec:veribench}.

\item As reward model is a important component for RL, we optimize the preference data generation with reward-tree construction, trace-back based rescore and preference data pair selection, followed by reward model training. details are given in \cref{sec:reward-model}.


\item To further improve the RL training of target LLM for stability and generalization, we first introduce a data taxonomy for training samples together with a novel Sample-balanced Weighting (SbW) reinforce++ algorithm (\cref{sec:SbW-Reinforce++}).

\end{itemize}

\section{Background and Related Work}

\subsection{LLM-based Verilog Code Generation}


Existing LLM-based approaches for Verilog code generation have primarily relied on techniques such as SFT and prompt engineering. For example, \eg, RTLCoder \cite{liu2024rtlcoder}, OriGen \cite{2024origen}, AutoVCoder \cite{autocoder2024}, HaVen \cite{yang2025haven} and CraftRTL \cite{liu2024craftrtl}, attempt to use large-scale synthetic or curated Verilog datasets during training. 
These methods allow models to learn specific patterns and syntactic structures by exposing them to vast amounts of data, which in turn leads to the generation of syntactically correct and functionally valid Verilog code in many cases.

However, these approaches also have notable drawbacks. 
Specifically, \emph{SFT-based methods tend to favor memorization over true understanding}, \ie, models perform well on known patterns from the training data but struggle to generalize to more complex or novel design challenges\cite{chu2025sft}. 
This overfitting to the training data hinders their ability to innovate or adapt when presented with new scenarios—a shortcoming that is particularly problematic in hardware design, where \emph{the diversity and intricacy of problems demand a deeper comprehension of underlying principles rather than a mere replication of memorized examples}.



\subsection{RL-based LLM Learning Method}


In contrast to SFT, RL based training methods \cite{yang2024acecode} \cite{zhang2024o1} empower LLMs with better reasoning capability, by learning through interaction and reward feedback.
One distinguished work is DeepSeek-R1 \cite{guo2025deepseek}, which employs RL to optimize model training, achieving advanced performance on multiple reasoning benchmarks. 
As illustrated in \cref{fig:framework-overview}(top), the training flow of DeepSeek-R1 is structured into four key stages:

\textbf{Stage-1:} 
To enhance the reasoning ability of the model and stabilize the early phase of RL, LLM (\ie, DeepSeek-V3-base) is trained by a cold-start SFT. 
The training data consists of several thousand high-quality chain-of-thought (CoT) \cite{wei2022chain} examples, each manually double-checked for readability. These samples follow the format \texttt{|special\_token| <reasoning process> |special\_token| <summary>}.

\textbf{Stage-2:} Building on the LLM trained by Stage-1, the model undergoes an RL process via group relative policy optimization (GRPO) \cite{shao2024deepseekmath} on reasoning tasks (\eg, math problems), where reward signal for RL is rule-based.
Its objective is to induce emergent and generalizable reasoning capabilities. 
Furthermore, the LLM generated in this stage is denoted as checkpoint\#1, which is used to conduct rejection sampling. Rejection sampling is a technique used to produce samples with high probability and independence from complex probability distributions\cite{liu2024statistical}. We conduct rejection sampling with the LLM checkpoint-1 by generating long CoT data and accepting some of them with high reward. Finally, Deepseek R1 generate 600K high-quality CoT samples by rejection sampling \cite{gilks1992adaptive}.

\textbf{Stage-3:} The base model (DeepSeek-V3-base) is fine-tuned again using the 600K reasoning samples generated in Stage-2, along with an additional 200K non-reasoning examples, resulting in a total of 800K training samples. This stage aims to improve the model's performance on both reasoning and non-reasoning tasks. The resulting model is referred to as Checkpoint \#2.

\textbf{Stage-4:} LLM is further trained using RL on full-scenario data starting from Checkpoint \#2. This training is designed to align the model’s outputs more closely with human preferences while reducing the risk of generating toxic or harmful content. The outcome of this stage is \textit{DeepSeek-R1}. 

\section{Framework of \sysname}
\label{sec:framework}
\subsection{Framework Overview}
\label{subsec:framework-overview}

\begin{figure*}[t]
    \centering
    \includegraphics[width=1\linewidth]{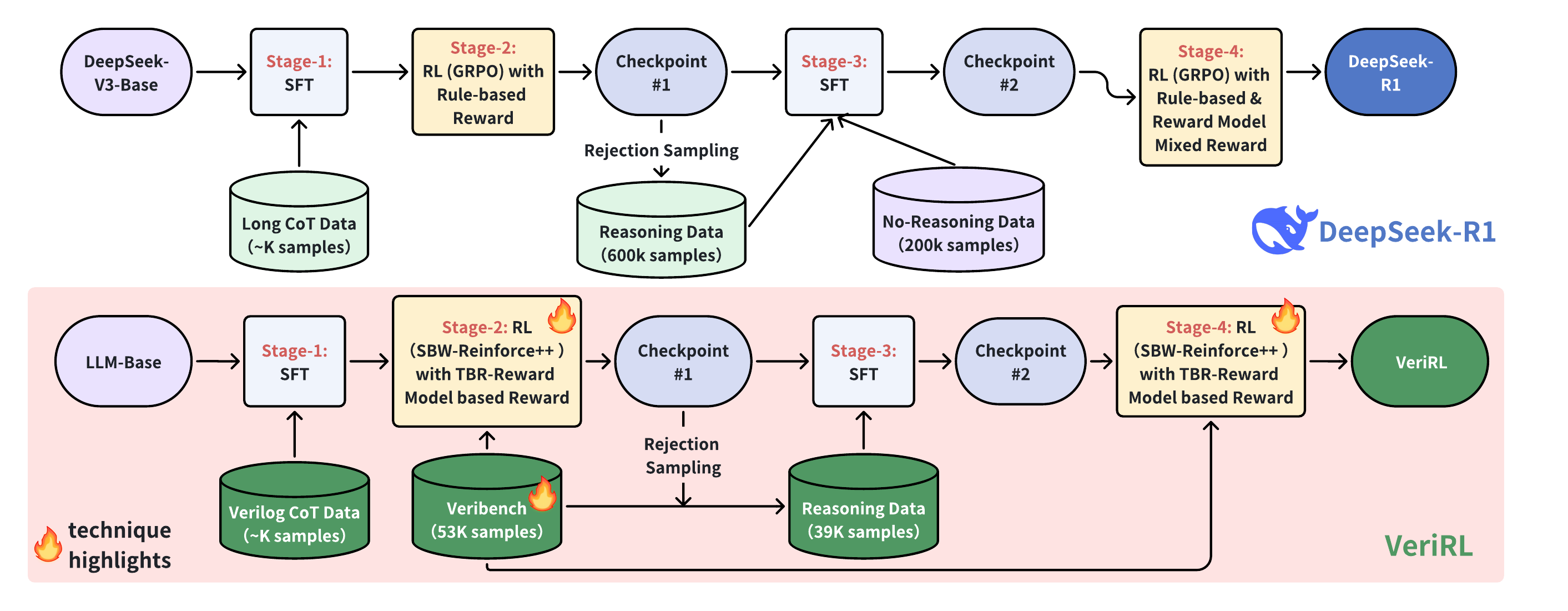}
    \caption{\textbf{Framework Overview of DeepSeek-R1 (Top) and Our \sysname (Bottom)}.  
    }
    \label{fig:framework-overview}
\end{figure*}


\textit{
Inspired by the success of DeepSeek-R1, we inherit and modify its training methodology to address the unique challenges of Verilog code generation.
Our proposed framework, \sysname, extends the four-stage RL-based training paradigm with domain-specific enhancements, which is detailed as:}

\textbf{Stage-1:} We train a LLM base model (default: 7B Qwen2.5-Coder) through SFT using a cold-start setting. 
The training data consists of several thousand high-quality CoT examples curated from GPT-4o-mini and manually verified for clarity and correctness. Each example follows the format of \texttt{<REASON>reasoning process</REASON> <SOLUTION>Code</SOLUTION>}.

\textbf{Stage-2:} Based on the model obtained in Stage-1, we perform RL using the classic Reinforce++ algorithm \cite{hu2025reinforce++}, enhanced with SbW(\cref{sec:SbW-Reinforce++}), tailored for Verilog generation. Unlike traditional rule-based reward generation methods (\eg, using testbenches to obtain the pass rates), we adopt a reward model (\cref{sec:reward-model}) that directly scores the functional correctness of the generated code. 
The resulting model from this RL process is denoted as Checkpoint \#1. We then apply rejection sampling with this checkpoint to generate 39K high-quality CoT samples (with pass rates exceeding 0.8), which are subsequently used to further improve the model’s general reasoning capabilities.

\textbf{Stage-3:} Another round of SFT using the same LLM, now with the expanded set of 39K high-quality reasoning samples. This phase strengthens the model's reasoning foundation and prepares it for stable RL training. The resulting model from this stage is denoted as Checkpoint \#2.

\textbf{Stage-4:} Ultimately, we apply RL again using Checkpoint \#2. This time, we adopt a SbW strategy, where gradient contributions are dynamically adjusted based on sample difficulty. This approach reduces training variance, mitigates forgetting, and improves overall RL stability. The final model obtained from this process is referred to as \sysname.

While \textsc{VeriRL} inherits the general training pipeline of DeepSeek-R1, it diverges in two key aspects: the \emph{design of the reward model} and the \emph{enhanced reinforcement learning algorithm}. These modifications are motivated by the unique challenges of Verilog code generation, which we elaborate on in the following section.

\subsection{Challenge 1: Low-Quality Reward Model Generate Sparse Reward Making RL-based Training Infeasible}

In \cref{fig:framework-overview}, a key challenge in applying RL to Verilog code generation lies in constructing a reliable reward model. Unlike natural language tasks, correctness in Verilog depends on simulation-based feedback, which is often sparse and brittle, making direct reward learning unstable.

To address this, we train a reward model using preference supervision based on pass rates. Specifically, we sample candidate programs from an open-source LLM (e.g., CodeQwen2.5) over VeriBench-53K and construct preference pairs $(x, [y^+, y^-])$ by comparing their simulation pass rates. However, LLMs often produce invalid or trivial code due to limited Verilog knowledge. Moreover, we observe that some generated programs, while containing minor syntax errors, exhibit logically correct behavior and have the potential to be refined into valid solutions. Discarding these candidates prematurely would result in a loss of valuable training signals. Relying solely on top-1 outputs further exacerbates this issue, leading to low-quality and uninformative preference data.

To mitigate this, we propose a \textbf{Trace-back based Rescore (TbR)} mechanism that recursively expands failed or low-reward programs by tracing intermediate reasoning steps and resampling alternatives. By selectively preserving and refining semantically meaningful but syntactically flawed outputs, TbR improves both the diversity and correctness of sampled programs, enabling more meaningful comparisons. It enhances the quality of the preference dataset without requiring additional model training and provides structured, simulation-grounded reward signals tailored for hardware domains. A concrete template used to instruct the TbR process is provided in Appendix~\ref{app:tbr-template}.

\subsection{Challenge 2: Instability in RL Training Caused by Knowledge Forgetting and Noisy Rewards}
\label{subsec:challenge-2}

Another major challenge in applying RL to Verilog code generation lies in the instability caused by imbalanced and noisy reward signals. Unlike natural language tasks, Verilog generation yields sparse correct outputs, and general-purpose LLMs often lack domain-specific knowledge, resulting in a large number of low- or zero-reward samples. This skews learning toward frequent but uninformative patterns, causing overfitting and catastrophic forgetting.

To address this, we propose a \textbf{Sample-balanced Weighting (SbW)} strategy that emphasizes rare but high-quality samples by scaling their gradient contributions based on both reward value and sampling frequency. This approach suppresses the dominance of noisy rollouts and promotes stable, generalizable policy learning---critical for domains like Verilog where supervision is sparse and evaluation costly.



\section{VeriBench-53K Generation}
\label{sec:veribench}

\begin{figure}
    \centering
    \includegraphics[width=1\linewidth]{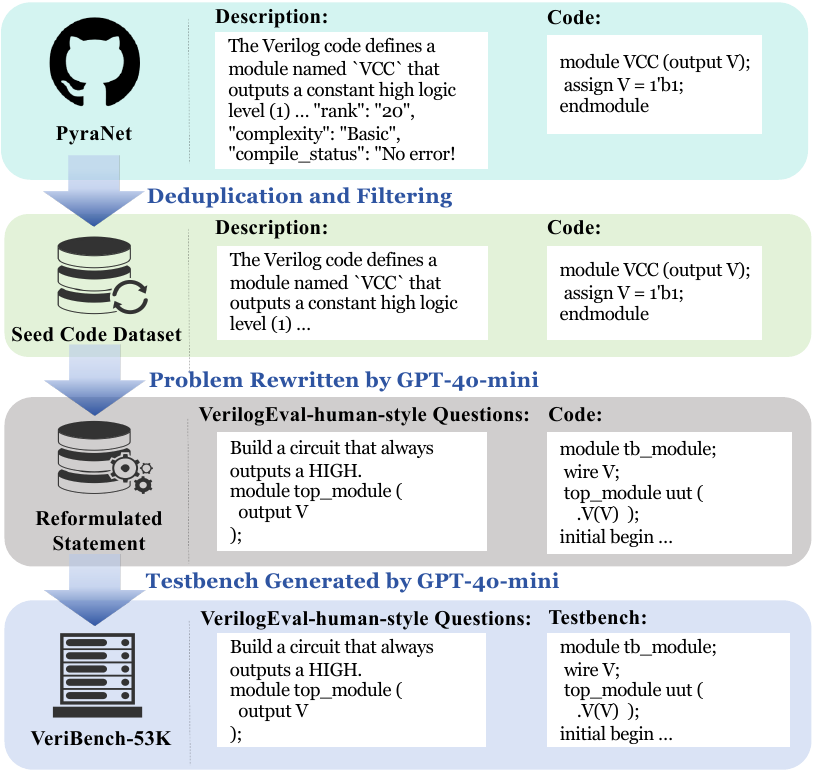}
    \caption{Overview of the dataset construction pipeline for Veribench-53K.}
    \label{fig:veribench-generation-flow}
\end{figure}

To support reward learning in our framework, particularly during the RL phases in Stage-2 and Stage-4, we construct \emph{Veribench-53K}, a large-scale dataset of Verilog problems paired with executable testbenches. This dataset addresses the critical challenge of obtaining reliable reward signals for Verilog code generation. Unlike mathematical tasks, where ground-truth answers can be verified through rule-based string matching or large-scale human annotations, Verilog code correctness must be validated through simulation. 
Evaluating whether a generated hardware module functions correctly under varying conditions requires executing testbenches and measuring their pass rates, a process that is computationally intensive and cannot be reduced to simple token-level heuristics. 
Moreover, existing Verilog corpora often lack verification artifacts or contain unvalidated examples, limiting their utility for fine-grained reward modeling.

As illustrated in \cref{fig:veribench-generation-flow}, Veribench-53K is generated through a three-step augmentation pipeline, which consists of three steps:


\textbf{(1) Deduplication and Filtering:} We begin with the open-source dataset \textit{PyraNet} \cite{nadimi2024pyranet}, a large-scale hierarchical repository containing over 700K Verilog design problems with functional descriptions. 
First, we apply MinHash-based deduplication with a Jaccard similarity threshold of 0.9 to remove near-duplicate entries. Next, we filter out all examples that fail to compile or contain unresolved dependencies, ensuring that each code snippet is syntactically valid and self-contained. Finally, we further filter by quality and difficulty by retaining only entries with a ranking score of at least 12 (as evaluated by GPT-4o-mini) and discarding entries labeled as \emph{Basic} while keeping those marked as Intermediate, Advanced, or Expert.

\textbf{(2) Problem Rewriting:} We observe that the original problem descriptions in PyraNet often resemble functional explanations rather than well-defined problem statements suitable for testbench-based evaluation. Therefore, we employ GPT-4o-mini to reformulate each functional description into a structured, VerilogEval-human-style problem statement \cite{liu2023verilogeval,pinckney2024revisiting} that explicitly specifies input/output ports, timing constraints, and behavioral requirements.

\textbf{(3) Testbench Generation:} For each curated problem, we generate a diverse set of testbenches. Specifically, GPT-4o-mini produces approximately 5--10 distinct testbenches for each problem, each designed to validate different aspects of the hardware behavior, including timing control, edge sensitivity, and functional correctness. This step provides robust, simulation-based evaluation signals that are essential for downstream reward model training. A concrete example of the prompt used to instruct GPT-4o-mini to generate testbenches is provided in Appendix~\ref{app:testbench-prompt}.


The curated Veribench-53K consists of high-quality Verilog problems with structured statements and richly annotated testbenches. Combined with the hierarchical metadata (\eg, ranking and complexity annotations) obtained from PyraNet, this dataset facilitates the training of reliable reward models for RL-based Verilog synthesis and verification. 

\section{Reward Model Construction}
\label{sec:reward-model}



\subsection{Preference Data for Reward Model Training}
In this work, we leverage the contrastive learning to train the reward model. Thus, it is necessary to construct data pairs in the form of (question, [positive answer, negative answer]) from the Veribench-53K dataset.
Given a question $q$ (\ie, VerilogEval-human-style question in \cref{sec:veribench}), we leverage a pretrained LLM (\eg, CodeQwen2.5) to generate $n$ designs ($n=10$ by default). Then, since each question owns 5-10 testbenches, we utilize those testbenches to get the ground-truth reward $r_i \in [0,1]$ (\ie, pass rate) of $i$-th answer ($i \in \{1,\cdots,n\}$). 
Categorizing answers as positive or negative is based on $r_i$, but we observe many answers come with $r_i=0$ or compilation failure.
As the countermeasure, we propose the tree-based CoT to enhance the preference data generation, including \emph{reward-tree construction} and \emph{trace-back based rescore}:

\begin{figure*}[t]
    \centering
    \includegraphics[width=\linewidth]{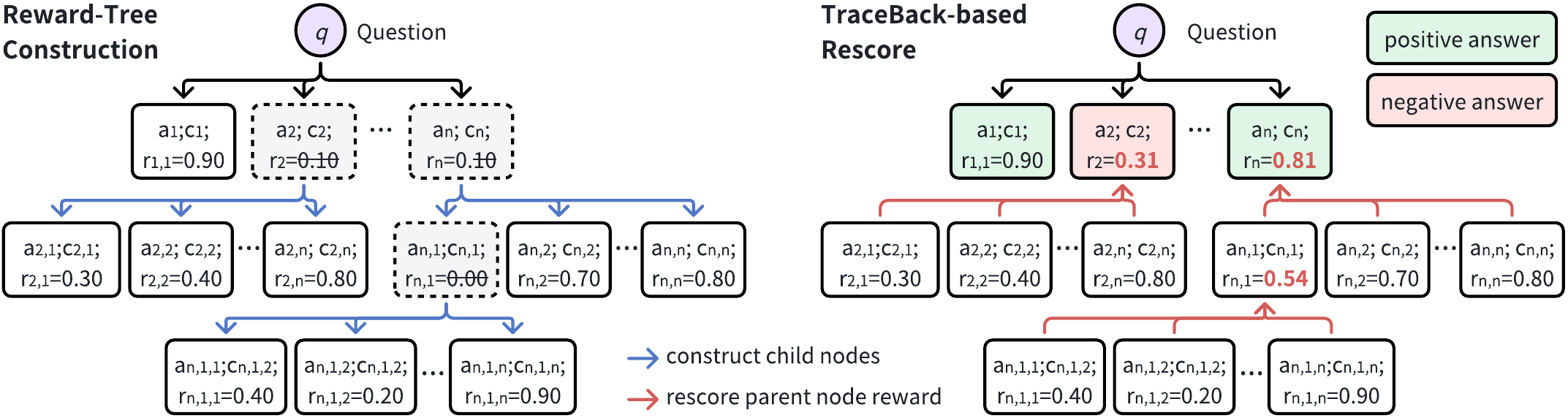}
    \caption{Reward tree construction with trace-back based rescore (TbR).}
    \label{fig:tree-COT}
\end{figure*}

\subsubsection{Reward-Tree Construction}
As shown in the left part of \cref{fig:tree-COT}, for the given question $q$, CodeQwen2.5 generates $n=10$ answers ($a_i$) and rewards ($r_i$) , as the first level of the reward tree.
Then, we set a threshold $0.2$ whose $r_i<0.2$ is flagged for regeneration.
Each flagged solution undergoes \emph{up to two rounds of self-reflection}. 
In every round, CodeQwen2.5 incorporates additional signals $c$, \eg compilation errors or testbench failure feedback, into its new reasoning process. 
This recursive refinement forms a tree structure where each node corresponds to a model-generated response, and edges represent the reflection path.

\subsubsection{Traceback based Rescore}
Once the reward-tree is constructed, we perform traceback based rescore (TbR) to update the rewards of root-level answers, as highlighted in \cref{fig:tree-COT}.
To rescore the nodes in this tree, we evaluate each response using the testbench. For any internal node (\ie, a parent response that triggered reflection), we replace its original score with the average reward of its child nodes, effectively backpropagating improvements made through self-reflection. This trace-back scoring avoids prematurely penalizing promising solutions due to superficial errors.

\subsubsection{Preference Data Pair Selection}
To make the positive and negative answers differ from each other significantly, for $i$-th and $j$-th answers ($i,j\in \{1,2,\cdots,n\}$ and $i \neq j$), they could be a preference data pair if satisfying the condition of:
\begin{equation}
\label{eqt:preference-pair-selection}
\1_{(r_i > r_j + 0.4) \land (r_i > 0.8) \land (r_j > 0)}
\end{equation}
$\1_\mathrm{condition}$ is 1 if the condition is true, 0 otherwise.
In this formulation, the margin \( r_i > r_j + 0.4 \) serves to ensure a substantial and meaningful performance gap between the positive sample and negative sample. This prevents the reward model from being trained on noisy or weak supervision signals, where the pass rates are too close to distinguish reliably. Additionally, we require the positive answer to have a pass rate greater than 0.8, thereby more likely representing a correct implementation.

\subsection{Reward Model Training}

Subsequently, we train the reward model $R_\phi$ ($\phi$ represents the model parameters) by performing full-parameter fine-tuning on an instruction-encoding model (\eg, CodeQwen2.5). Specifically, we extract the final hidden representation of the last token and feed it into a linear head to produce a scalar output, which is optimized using the loss function defined in \cref{reward_loss}.
Assume $(\hat{r}_1,\cdots,\hat{r}_n)$ are the rescored rewards. When training the reward model $R_{\phi}$ using Bradley-Terry loss \cite{bradley1952rank}, the loss for each preference pair $a_{i}$ and $a_{j}$ is defined as:
\begin{equation}
    \1_{\hat{r}_i>\hat{r}_j} 
    \log\sigma(R_{\phi}(q,a_{i})-R_{\phi}(q,a_{j})),
\end{equation}
The loss function for the reward model training is:
\begin{equation}\label{reward_loss}
    \mathcal{L}(\phi)=-\frac{1}{m}\sum_{r_{i}>r_{j}} \mathcal{L}_{\phi}(x,\hat{r}_i,\hat{r}_j),
\end{equation}
where $m$ is the number of preference pairs. This Loss means the reward model is trained to assign higher values to preferred respond and lower value to non-preferred ones, maximizing the difference between these ratings.

\section{SbW-based Reinforce++ for LLM Learning}
\label{sec:SbW-Reinforce++}

\begin{figure}
    \centering
    \includegraphics[width=1\linewidth]{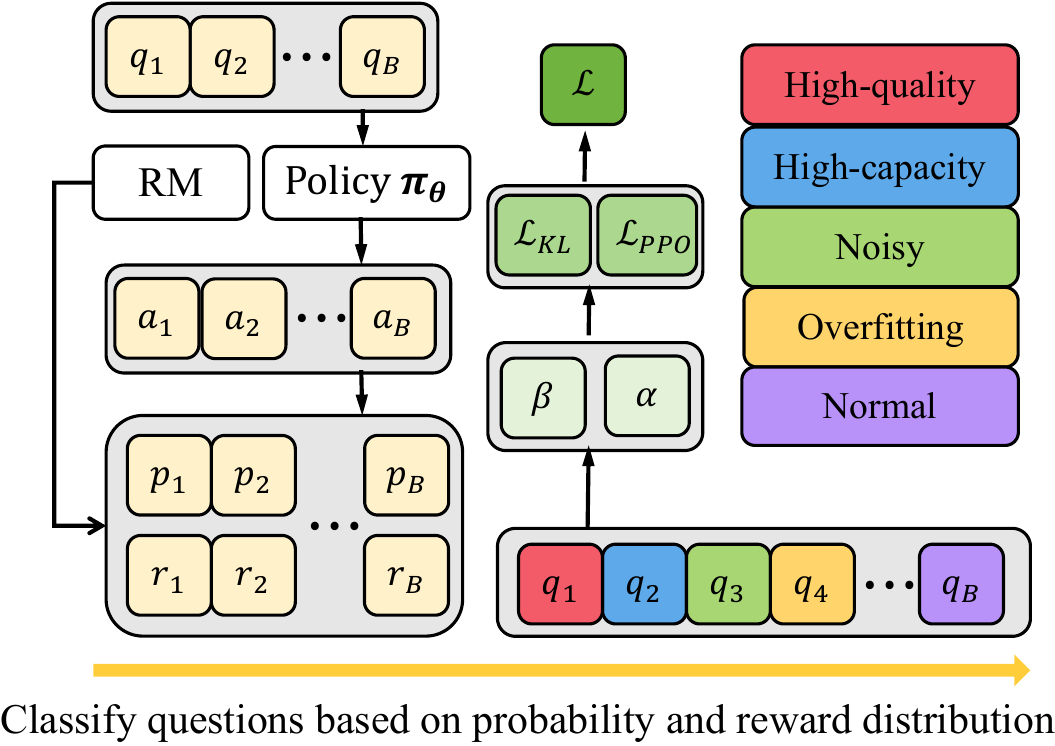}
    \caption{Overview of the SbW-based Reinforce++.}
    \label{fig:SbW-reinforce}
\end{figure}

As depicted in \cref{fig:SbW-reinforce}, our LLM used for Verilog code generation is the policy model (\ie, \sysname $\pi_\theta$) in RL parameterized by $\theta$. Given a batch of question $\{q_1,q_2,\cdots, q_B\}$, the policy model generates corresponding answers $\{a_1,a_2,\cdots,a_B\}$ with generation probabilities $\{p_1,p_2,\cdots,p_B\}$, where $p_i = P_{\pi_\theta}(q_i, a_i)$. Meanwhile, the reward model (RM) generated in \cref{sec:reward-model} is utilized to evaluate each generated answer with reward $\{r_1,r_2,\cdots, r_B\}$, where $r_i = R_\phi(q_i,a_i)$.
Based on the probability $p_i$ and reward $r_i$ of question-answer pair, we first divide the data into five categories \cref{subsec:data-taxonomy}, then assign specific weights to the policy loss \cref{subsec:sbw-reinforce}.

\subsection{Taxonomy of Data Entry}
\label{subsec:data-taxonomy}

To categorize the question-answer pair, we set a high threshold $G_{p}=\mu_{p}+\sigma_{p}$ and a low threshold $F_{p}=\mu_{p}-\sigma_{p}$ for probability, with a high threshold $G_{r}=\mu_{r}+\sigma_{r}$ and a low threshold $F_{r}=\mu_{r}-\sigma_{r}$ for reward. \cite{zhang2024cppo}

\begin{table}[t]
\centering
\footnotesize
\begin{tabular}{cccccc}
\toprule
Category & \multicolumn{2}{c}{Determining Condition} & \multicolumn{2}{c}{Weighting Strategy} \\
\midrule
High-quality & $P_{\pi_{\theta}}(q,a)\geq G_{p}$ & $R_\phi(q,a)\geq G_{r}$ &$\alpha\uparrow$ &$\beta\uparrow$\\
Overfitting & $P_{\pi_{\theta}}(q,a)\geq G_{p}$ & $R_\phi(q,a)\leq F_{r}$ &$\alpha\uparrow$ &$\beta\downarrow$\\
High-capacity & $P_{\pi_{\theta}}(q,a)\leq F_{p}$ & $R_\phi(q,a)\geq G_{r}$ &$\alpha\uparrow$ &$\beta\downarrow$\\
Noisy & $P_{\pi_{\theta}}(q,a)\leq F_{p}$ & $R_\phi(q,a)\leq F_{r}$ &$\alpha\downarrow$ &$\beta\downarrow$\\
Normal & \multicolumn{2}{c}{$P_{\pi_{\theta}}$ or $R_\phi$ not in above for region} &- &-\\
\bottomrule
\end{tabular}
\caption{Taxonomy of question-answer pair based on generation probability and reward evaluated by the reward model.}
\label{tab:taxonomy}
\end{table}

\subsection{Reinforce++ with Sample-balanced Weighting Strategy}
\label{subsec:sbw-reinforce}

To efficiently update the policy model, we introduce the loss composed of \emph{Kullback–Leibler divergence} term to avoid catastrophic forgetting and \emph{PPO loss} term to learn new knowledge. Correspondingly, we define advantage weighting term $\alpha(q,a)$ to encourage learning and $\beta(q,a)$ to penalize deviation from the old policy. 
\begin{equation}
    \begin{aligned}
        &\min_\theta \mathcal{L}(q|\theta)= \beta(q,a)KL(p_{\pi_{\theta}}(a|q)|p_{\pi_{old}}(a|q))
        \\&- \alpha(q,a)\min [r(\theta)A,\textrm{clip}(r(\theta),1-\epsilon,1+\epsilon)A] 
    \end{aligned}
\end{equation}


The advantage function $A$, which indicates how much better an action is compared to the average. The probability ratio $r(\theta)=\frac{p_{\pi_{\theta}}(a|q)}{p_{\pi_{old}}(a|q)}$, which adjusts the advantage based on the change in policy. The clipped objective $\textrm{clip}(r(\theta),1-\epsilon,1+\epsilon)$, which ensures that the policy update does not drastically change the action probabilities (preventing large updates). $\alpha$ encourages the model to think deeply about new knowledge, and $\beta$ constrains the model updates. 
The value assignment of $\alpha$ and $\beta$ depends on the data sample taxonomy given in \cref{tab:taxonomy}. Their expression can be written as:
\begin{equation}
    \begin{aligned}
        \alpha(q,a)&=\max\{\frac{P_{\pi_{\theta}}(q,a)}{\mu[P_{\pi_{\theta}}(q,a)]},\frac{R_\phi(q,a)}{\mu[R_\phi(q,a)]}\}\\
        \beta(q,a)&=\min\{\frac{P_{\pi_{\theta}}(q,a)}{\mu[P_{\pi_{\theta}}(q,a)]},\frac{R_\phi(q,a)}{\mu[R_\phi(q,a)]}\}
    \end{aligned}
\end{equation}
Normalization via the mean $\mu$ can mitigate the impact of extreme values, ensuring robust weighting strategy. In addition, the use of $\max$ and $\min$ dynamically adjusts $\alpha$ and $\beta$, enabling adaptive weighting allocation based on varying conditions.


\section{EXPERIMENTAL SETTINGS}

\subsection{Foundation Models}
Following previous works, we adopt several state-of-the-art large language models for code generation as our foundation models. Specifically, we use Qwen2.5-Coder-7B-Instruct as the base model for training the reward model. For reinforcement learning, we leverage Deepseek-Coder-6.7B-Instruct and Qwen2.5-Coder-7B-Instruct. Qwen2.5-Coder-7B-Instruct belongs to the Qwen2.5-Coder family, which represents one of the most advanced code-focused LLM series currently available.

\subsection{Datasets}
\subsubsection{Veribench53K}
We first construct Veribench-53K, a curated dataset extracted from the large-scale PyraNet corpus of 700K Verilog problems [16].
We apply MinHash-based deduplication, compilation filtering,
and quality screening (PyraNet’s original score and complexity
labels), preserving only Intermediate to Expert problems with
a score $\geq$ 12. Each sample is further augmented through: (i)
structured VerilogEval-human-style prompt [17] reformulation to clarify design intent, and (ii) testbench synthesis, generating
5-10 diverse testbenches per problem covering edge sensitivity,
timing, and functionality
\subsubsection{Reward model}
To train the reward model, we construct a preference dataset from the Veribench-53K benchmark. Each data point consists of a question and a pair of model-generated Verilog code implementations labeled as (positive, negative). For each question, we sample 10 candidate solutions using a pretrained LLM (e.g., CodeQwen2.5), and evaluate them against 5–10 testbenches to obtain ground-truth rewards based on functional correctness (i.e., pass rates). Answers are categorized as positive or negative depending on the reward gap, with additional filtering criteria to ensure significant quality differences. To handle low-quality or failing outputs (e.g., compilation errors), we apply a tree-based refinement strategy that allows the model to reflect and regenerate improved responses. After applying these procedures, we obtain approximately 240,000 preference pairs, which serve as supervision for training the reward model.
\subsubsection{Reinforcement Learning}
We train the model on a subset of the Veribench-53K dataset, which consists of the bottom 40\% of problems ranked by average pass rate and variance. This subset is chosen to ensure both sufficient difficulty and diverse program behaviors.
\subsection{Model Training}
\subsubsection{Reward model Training}
The reward model is trained using LlamaFactory Framework \cite{zheng2024llamafactory}, with full fine-tuning enabled via DeepSpeed Framework Stage 3. Reward training is conducted for 1 epoch using a cosine learning rate schedule, starting at a learning rate of $1 \times 10^{-5}$ and a warmup ratio of 0.1, allowing a smooth ramp-up during the initial phase. The batch size is set to 128. We enable bf16 precision to reduce memory usage without compromising model performance. Training takes approximately 24 hours on 8$\times$A100 GPUs.
\subsubsection{Reinforcement Learning Training}
We use Qwen2.5-Coder-7B-Instruct and DeepSeek-Coder as the policy model for RL training, with the reward signals provided by a previously trained reward model. The RL fine-tuning is implemented using the OpenRLHF \cite{hu2024openrlhf} framework. We adopt the Reinforcement++ algorithm as a replacement for PPO to improve training efficiency while avoiding the need for a value model. This algorithm has also been shown to be more stable than both PPO and GRPO.

To mitigate overfitting and catastrophic forgetting during the RL process, we introduce a \textbf{SbW} strategy. This strategy samples problems based on their probability and reward, ensuring that the RL agent does not over-adapt to frequent or trivial patterns.

We train the model on a subset of the \textbf{Veribench-53K} dataset, which consists of the bottom 40\% of problems ranked by average pass rate and variance. This subset is chosen to ensure both sufficient difficulty and diverse program behaviors. For training hyperparameters, we set the rollout batch size to 256 and sample 4 solutions per question. The training batch size is set to 128, with a learning rate of \(5 \times 10^{-7}\). All models are trained for one episode and the training process takes approximately 6 hours on 8$\times$A100 GPUs.
\subsection{Model Inference}
During inference, only the design specification x is available for the model. For our experiments, we used vLLM with specific configurations for the inference engine. The engine operates with the bf16 type and utilizes tensor parallelism across four devices, while maintaining a maximum token limit of 4096. We configure the sampling parameters with top p at 0.95 and top k at 50. Following the methodology established in previous work, we reported optimal results in three temperature: 0.2, 0.5, and 0.8.
\subsection{Benchmarks}
We evaluated our models on two representative benchmarks for Verilog code generation: RTLLM v1.1 and VerilogEval-human. Both benchmarks require LLMs to generate RTL designs based on natural language specifications. RTLLM v1.1 consists of 29 RTL design tasks, including 11 arithmetic designs and 18 logic designs. For VerilogEval, we only consider the VerilogEval-human subset, which contains 156 manually crafted problems. We exclude VerilogEval-machine from our evaluation, as it was generated by GPT-based models and is known to deviate from realistic industrial design requirements. Notably, VerilogEval v2 has also dropped the machine-generated subset, further indicating its limited practical relevance.

In addition, we perform evaluations on RTLLM v2, which extends RTLLM v1.1 to 50 RTL tasks across four categories: Arithmetic, Memory, Control, and Miscellaneous. We also evaluate on VerilogEval v2, which builds upon the VerilogEval-human set by adopting a chatbot-style interface to better simulate real-world interactions.
\subsection{Metrics}
We evaluate the performance of Verilog code generation models primarily from the perspective of \emph{functional correctness}, using the widely adopted \textit{pass@k} metric~\cite{chen2021evaluatinglargelanguagemodels}, which estimates the proportion of problems for which at least one of \(k\) generated solutions passes functional verification (\(k \in \{1, 5\}\)). The metric is defined as:
\begin{equation}
\text{pass}@k \coloneqq \mathbb{E} \left[ 1 - \frac{\binom{n - c}{k}}{\binom{n}{k}} \right],
\end{equation}
where \(n \geq k\) is the number of sampled outputs per problem and \(c\) is the number of correct outputs among them. Following the experimental setting in VerilogEval~\cite{liu2023verilogeval}, we set \(n = 20\) in all our experiments.

We focus exclusively on functional correctness (pass@1 and pass@5) in our evaluation. Although syntax correctness was previously a concern in early LLM-based code generation, modern LLMs have demonstrated consistently high syntax validity for Verilog code, making syntax errors relatively rare. As such, we regard syntax correctness as no longer the primary bottleneck, and instead emphasize functional correctness as the more meaningful and challenging evaluation criterion.

\section{RESULTS AND ANALYSIS}

\begin{table*}[htbp]
\centering
\setlength{\tabcolsep}{3pt}
\caption{Comparison of Verilog models on VerilogEval and RTLLM v1.1.}\label{main}
\begin{tabular}{llccccccc}
\toprule
\multirow{2}{*}{\textbf{Source}} & \multirow{2}{*}{\textbf{Name}} & \multirow{2}{*}{\textbf{Open source}} & \multirow{2}{*}{\textbf{ModelSize}} &
\multicolumn{2}{c}{\textbf{VerilogEval-human (\%)}} & 
\multicolumn{2}{c}{\textbf{RTLLM v1.1 (\%)}} \\
\cmidrule(lr){5-6} \cmidrule(lr){7-8}
 & & & & \makebox[1.5cm][c]{pass@1} &
\makebox[1.5cm][c]{pass@5} & \makebox[1.5cm][c]{pass@1} & \makebox[1.5cm][c]{pass@5} \\
\midrule
\multirow{5}{*}{General LLM} 
 & GPT-3.5 & $\Box$ & -- & 26.7 & 45.8 & 29.1 & 37.9 \\
 & GPT-4o &$ \Box$ & -- & 57.1 & 63.9 & 47.9 & 58.0 \\
 & Starcoder \cite{li2023starcoder} & $\CheckedBox$ & 15B & 18.1 & 26.6 & 18.3 & 27.6 \\
 & CodeLlama \cite{roziere2023code} & $\CheckedBox$ & 7B & 18.2 & 22.7 & 17.9 & 31.0 \\
 & DeepSeek-Coder \cite{guo2024deepseekcoderlargelanguagemodel} & $\CheckedBox$ & 6.7B & 30.2 & 33.9 & 23.1 & 29.3 \\
 & CodeQwen1.5 \cite{bai2023qwen} & $\CheckedBox$ & 7B & 22.5 & 26.1 & 24.1 & 34.2\\
 & CodeQwen2.5 \cite{hui2024qwen2} & $\CheckedBox$ & 7B & 31.0 & 43.0 & 28.6 & 43.7\\
\midrule
\multirow{11}{*}{Verilog-Specific Models}
 & ChipNeMo \cite{liu2024chipnemo} & $\Box$ & 13B & 22.4 & -- & -- & -- \\
 & Thakur et al. \cite{thakur2023benchmarking} & $\CheckedBox$ & 16B & 30.3 & 43.9 & -- & 24.1 \\
 & RTLCoder-Mistral \cite{liu2024rtlcoder} & $\CheckedBox$ & 7B & 36.7 & 45.5 & 24.5  & 37.3 \\
 & RTLCoder-DeepSeek-Coder \cite{liu2024rtlcoder} & $\CheckedBox$ & 6.7B & 41.6 & 50.1 & 35.8 & 40.3 \\
 & BetterV-CodeLlama \cite{pei2024betterv} & $\Box$ & 7B & 40.9 & 50.0 & -- & -- \\
 & BetterV-DeepSeek-Coder \cite{pei2024betterv} & $\Box$ & 6.7B & 45.9 & 53.3 & -- & -- \\
 & BetterV-CodeQwen1.5 \cite{pei2024betterv} & $\Box$ & 7B & 46.1 & 53.7 & -- & -- \\
 & AutoVCoder-CodeLlama \cite{autocoder2024} & $\Box$ & 7B & 44.5 & 52.8 & -- & 48.3 \\
 & AutoVCoder-DeepSeek-Coder \cite{autocoder2024} & $\Box$ & 6.7B & 46.9 & 53.7 & -- & 51.7 \\
 & AutoVCoder-CodeQwen1.5 \cite{autocoder2024} & $\Box$ & 7B & 48.5 & 55.9 & -- & 51.7 \\
 & OriGen-DeepSeek-Coder-7B-v$1.5^{\star}$  \cite{2024origen} & $\CheckedBox$ & 7B & 54.4 & 60.1 & -- & {\cellcolor[HTML]{CD7F32}65.5}\\
 & HaVen-CodeQwen1.5 \cite{yang2025haven} & $\CheckedBox$ & 7B & 61.1 & 64.8 & 50.3 & 62.2 \\
 & CraftRTL-CodeLlama \cite{liu2024craftrtl} & $\Box$ & 7B & 63.1 & 67.8 & 42.6 & 52.9 \\
 & CraftRTL-DeepSeek-Coder \cite{liu2024craftrtl} & $\Box$ & 6.7B & 65.4 & 70.0 &
 {\cellcolor[HTML]{CD7F32}53.1} & 58.8 \\
 & CraftRTL-Starcoder2 \cite{liu2024craftrtl} & $\Box$ & 15B & {\cellcolor[HTML]{C0C0C0}68.0} & {\cellcolor[HTML]{CD7F32}72.4} & 49.0 &{\cellcolor[HTML]{C0C0C0}65.8} \\
\midrule

\multicolumn{2}{l}{\textbf{Ours: VeriRL-DeepSeek-Coder}} & $\CheckedBox$ & 6.7B & {\cellcolor[HTML]{CD7F32}68.7} & {\cellcolor[HTML]{C0C0C0}77.6} & 
{\cellcolor[HTML]{C0C0C0}55.3} & 65.4 \\

\multicolumn{2}{l}{\textbf{Ours: VeriRL-CodeQwen2.5 }} & $\CheckedBox$ & 7B & \textbf{\cellcolor[HTML]{FFD800}69.3} & \textbf{\cellcolor[HTML]{FFD800}78.1} &
\textbf{\cellcolor[HTML]{FFD800}58.2} & 
\textbf{\cellcolor[HTML]{FFD800}66.0} 
\\

\rowcolor{blue!10}
\bottomrule
\multicolumn{2}{l}{
$\dagger$: ranking of performance as \textcolor{gold}{\faTrophy} 1st, \textcolor{silver}{\faTrophy} 2nd, \textcolor{bronze}{\faTrophy} 3rd place; \hspace{1em}
}
\end{tabular}
\end{table*}

\begin{table*}[htbp]
\centering
\setlength{\tabcolsep}{3pt}
\caption{Comparison of Verilog models on VerilogEval v2 and RTLLM v2.}\label{main_v2}
\begin{tabular}{llccccccc}
\toprule
\multirow{2}{*}{\textbf{Source}} & \multirow{2}{*}{\textbf{Name}} & \multirow{2}{*}{\textbf{Open source}} & \multirow{2}{*}{\textbf{ModelSize}} &
\multicolumn{2}{c}{\textbf{VerilogEval v2 (\%)}} & 
\multicolumn{2}{c}{\textbf{RTLLM v2 (\%)}} \\
\cmidrule(lr){5-6} \cmidrule(lr){7-8}
 & & & & \makebox[1.5cm][c]{pass@1} &
\makebox[1.5cm][c]{pass@5} & \makebox[1.5cm][c]{pass@1} & \makebox[1.5cm][c]{pass@5} \\
\midrule
\multirow{5}{*}{General LLM} 
 & GPT-4o &$ \Box$ & -- & 56.5 & 65.1 & 47.9 & 58.0 \\
 & CodeLlama \cite{roziere2023code} & $\CheckedBox$ & 7B & 12.1 & 20.2 & 25.8 & 29.0 \\
 & DeepSeek-Coder \cite{guo2024deepseekcoderlargelanguagemodel} & $\CheckedBox$ & 6.7B & 22.6 & 29.6 & 26.5 & 36.3 \\
 & CodeQwen1.5 \cite{bai2023qwen} & $\CheckedBox$ & 7B & 19.8 & 24.3 & 25.8 & 29.0\\
 & CodeQwen2.5 \cite{hui2024qwen2} & $\CheckedBox$ & 7B & 34.4 & 43.8 & 40.5 & 49.8\\

\midrule
\multirow{4}{*}{Verilog-Specific Models}

 & RTLCoder-Mistral \cite{liu2024rtlcoder} & $\CheckedBox$ & 7B & 34.0 & 37.5 & 37.0  & 39.9 \\
 & RTLCoder-DeepSeek-Coder \cite{liu2024rtlcoder} & $\CheckedBox$ & 6.7B & 40.9 & 47.9 & 43.5 & 48.0 \\
 & OriGen-DeepSeek-Coder-7B-v$1.5^{\star}$  \cite{2024origen} & $\CheckedBox$ & 7B & 51.3 & 56.3 & 40.9 & 57.1\\
 & HaVen-CodeQwen1.5 \cite{yang2025haven} & $\CheckedBox$ & 7B & 53.1 & 61.8 & 49.2 & 61.7 \\
\midrule

\multicolumn{2}{l}{\textbf{Ours: VeriRL-DeepSeek-Coder}} & $\CheckedBox$ & 6.7B & {66.2} & 73.2 & 60.1 & 68.2 \\

\multicolumn{2}{l}{\textbf{Ours: VeriRL-CodeQwen2.5 }} & $\CheckedBox$ & 7B & \textbf{67.2} & \textbf{76.1} & \textbf{63.3} & 
\textbf{70.3} 
\\
\bottomrule
\multicolumn{2}{l}{
\hspace{1em}
}
\end{tabular}
\end{table*}
In this section, we present comprehensive experiments to evaluate the effectiveness of our reinforcement learning framework for Verilog code generation. We organize our evaluation into five parts: overall performance, ablation studies, detailed analysis of the Trace-back based Rescore (TbR) mechanism, the Sample-balanced Weighting (SbW) strategy, and generalization and robustness tests.
\subsection{Main Results}
Table \ref{main} presents a comprehensive comparison of our models, DeepSeek-Coder and CodeQwen2.5, with both general-purpose LLMs (e.g., GPT-4o, Starcoder) and Verilog-specific models (e.g., BetterVCoder, AutoVCoder) on two widely-used benchmarks: VerilogEval-Human and RTLLM v1.1. Our CodeQwen2.5 model consistently achieves state-of-the-art results across all evaluation metrics. Specifically, it attains 69.3\% pass@1 and 78.1\% pass@5 on VerilogEval-Human, outperforming all baseline models. On RTLLM v1.1, it achieves 58.2\% pass@1 and 66.0\% pass@5, again ranking first among all compared models.

Compared to other 7B-scale Verilog-specialized models, CodeQwen2.5 surpasses previously strong baselines such as HaVen-CodeQwen1.5 and CraftRTL-Deepseek-Coder, which previously held leading performance. Notably, our model outperforms these prior models, especially CraftRTL-DeepSeek-Coder by 3.9\% and 5.1\% in pass@1 on VerilogEval-Human and RTLLM v1.1, respectively. Even compared to the 15B CraftRTL-Starcoder2, CodeQwen2.5 still yields a 5.7\% improvement in pass@5 on VerilogEval-Human.

Table \ref{main_v2} further evaluates these models on the more challenging VerilogEval v2 and RTLLM v2 benchmarks. Again, our CodeQwen2.5 achieves the highest pass@1 and pass@5 scores, with 67.2\% / 76.1\% on VerilogEval v2 and 63.3\% / 70.3\% on RTLLM v2. These represent new state-of-the-art performances, exceeding prior models such as Haven-CodeQwen by 14.3\% on VerilogEval v2 pass@5, and 8.6\% on RTLLM v2 pass@5, respectively.
These results underscore the strength of our CodeQwen2.5 model in understanding, generating, and reasoning about Verilog, even when compared against larger-scale or domain-tuned alternatives.

\begin{table}[ht]
\centering
\caption{Ablation Study on VerilogEval v1 Benchmarks. We report pass@1 and pass@5. Each component incrementally improves performance.}
\label{tab:ablation}
\begin{tabular}{lcc}
\toprule
\textbf{Model Variant} & \textbf{pass@1 (\%)} & \textbf{pass@5 (\%)} \\
\midrule
Baseline (Reinforce++, no TbR/SbW)   & 53.2 & 61.5 \\
+ TbR                                & 65.3 & 73.3 \\
+ SbW                                & 64.7 & 69.2 \\
+ TbR + SbW (Ours)                   & \textbf{69.3} & \textbf{78.1} \\
\midrule
\end{tabular}
\end{table}

\subsection{Ablation Study}
To assess the individual contributions of each proposed component, we perform ablation studies on the VerilogEval v1 benchmark. Starting from a baseline model fine-tuned with our Reinforce++ framework but without the Trace-back based Rescore (TbR) or the Sample-balanced Weighting (SbW) strategy, we incrementally integrate each module and evaluate the resulting performance.

The results, summarized in Table    \ref{tab:ablation}, demonstrate that both TbR and SbW contribute substantially to model improvement. Specifically, adding TbR alone significantly boosts performance from 53.2\% to 65.3\% pass@1, showing that recovering and leveraging previously discarded, partially correct samples provides valuable learning signals. Similarly, introducing SbW independently improves the model to 64.7\% pass@1, validating its effectiveness in stabilizing training and reducing variance through adaptive sample weighting. When both mechanisms are combined, the model achieves its best performance, reaching 69.3\% pass@1 and 78.1\% pass@5, indicating a strong complementary effect between TbR and SbW.


\subsection{Analysis of Trace-back based Rescore Mechanism (TbR)}
\begin{figure}
    \centering
    \includegraphics[width=1\linewidth]{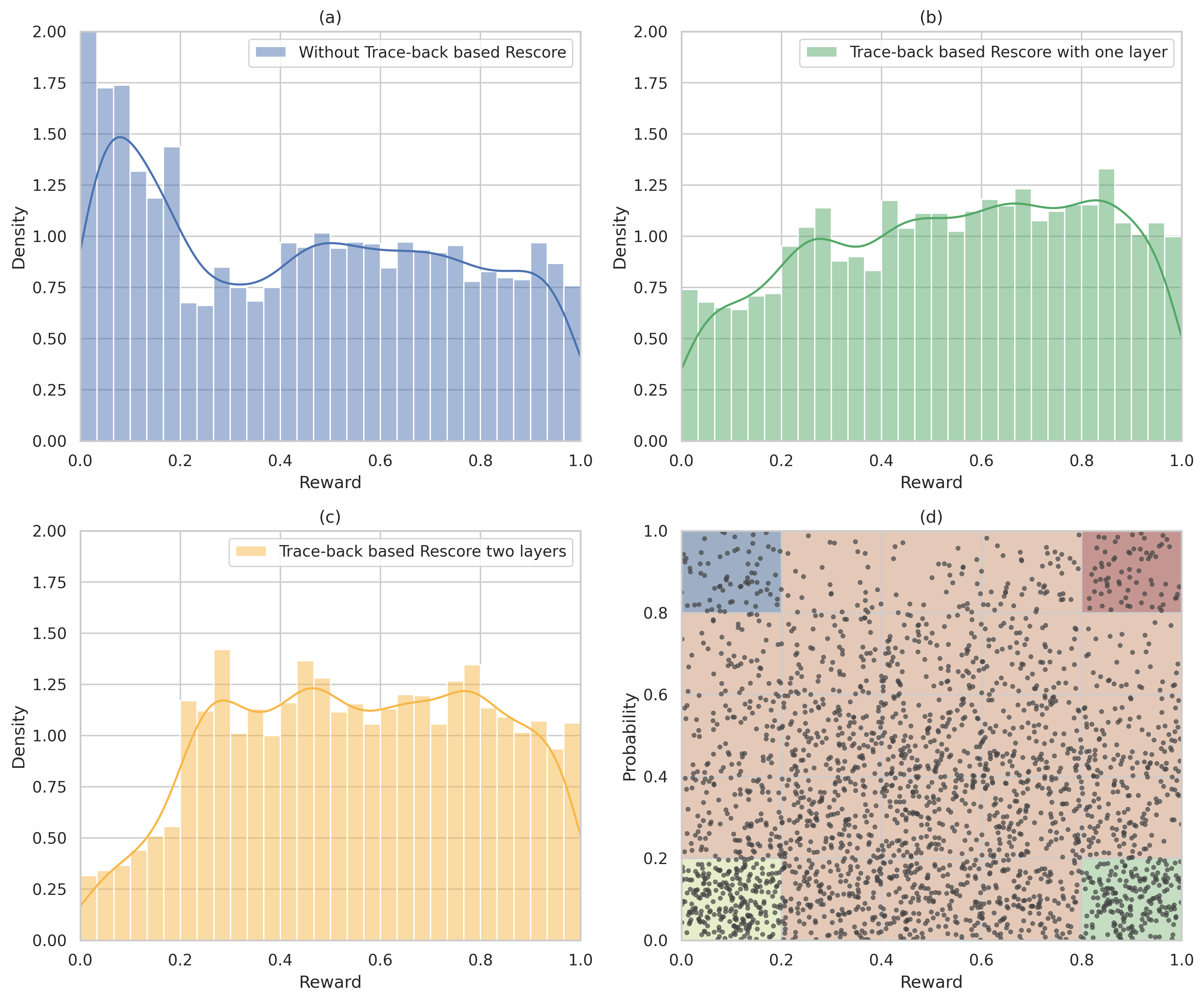}
    \caption{Reward-Density and Probability-Reward Scatter.}
    \label{all}
\end{figure}
We conduct in-depth experiments to assess the impact of the Trace-back based Rescore (TbR) mechanism on reward quality and final model performance. First, we analyze the distribution of reward values before and after applying TbR. As shown in Figure~\ref{all}(a-c), without TbR, the reward values are heavily skewed, with the majority of samples receiving zero or near-zero rewards. This is primarily due to the prevalence of syntactically invalid or functionally incorrect code among raw LLM outputs. After introducing TbR, the distribution becomes significantly smoother and more continuous, indicating a higher proportion of candidates receiving meaningful reward signals. This flattening effect improves the diversity and informativeness of the reward landscape, which is beneficial for both reward model training and reinforcement learning.
\begin{table}[t]
\centering
\footnotesize
\begin{tabular}{lccc}
\toprule
\textbf{Iteration Rounds} & 
\textbf{\# Pref. Pairs} &
\textbf{VerilogEval} & 
\textbf{RTLLM v1.1}\\
\midrule
w/o TbR  & 65k & 61.5 & 51.3 \\
w TbR Iterate 1 & 129k & 69.8 & 58.4 \\
w TbR Iterate 2 & 210k & 72.1 & 64.1 \\
w TbR Iterate 3 & 240k & 76.3 & 66.0 \\
\bottomrule
\end{tabular}
\caption{Impact of iterative TbR on reward data volume and model performance. Each iteration expands the preference dataset and improves final pass@1 accuracy on VerilogEval and pass@5 accuracy on RTLLM v1.1.}
\label{tab:tbrround}
\end{table}

We further investigate how iterative application of TbR influences the construction of the preference dataset and downstream performance. Table~\ref{tab:tbrround} reports the number of collected training samples and the corresponding pass@1 scores on VerilogEval as TbR is applied across multiple rounds. Without TbR, only 65k preference pairs can be extracted due to the low quality of initial generations, and the resulting model achieves a pass@1 of 61.5\%. After one round of TbR, the dataset expands to 129k samples, yielding a pass@1 of 69.8\%. Continuing this iterative process, the dataset grows to 210k and 240k samples after the second and third rounds, respectively, with the final model reaching a pass@1 of 76.3\%. These results highlight that each round of TbR not only increases the volume of usable training data but also leads to consistent improvements in model performance, demonstrating the effectiveness of this self-improving loop.

To further illustrate the effectiveness of the Trace-back based Rescore mechanism, we present a representative case study in  Appendix~\ref{case:sync_counter}. The initial candidate compiles successfully but exhibits incorrect reset behavior, achieving a pass rate of only 3/10. Specifically, the design implements an asynchronous reset by including posedge rst in the sensitivity list, which violates the intended synchronous semantics.

After applying TbR, the model corrects this subtle semantic error by removing posedge rst and moving the reset condition inside the posedge clk block. The revised code passes all test cases and is retained as a high-quality preference sample. This case highlights how TbR enables the recovery and refinement of otherwise discarded candidates, thereby enhancing dataset quality and improving downstream model performance.


\subsection{Analysis of Sample-balanced Weighting (SbW)}
To understand the impact of the Sample-balanced Weighting (SbW) strategy, we analyze its effect on training stability and reward learning. Specifically, we examine how SbW reshapes the reward-probability space and regulates the optimization dynamics.

In Fig \ref{all} (d), we visualize the joint distribution of generation probabilities and rewards, enabling a clear categorization of samples into five types. Notably, regions representing noisy, overfitting, and high-capacity samples illustrate imbalances that could destabilize RL training if treated uniformly. By assigning adaptive weights based on sample category, SbW strategy ensures stable gradient updates and effectively balances learning progress with knowledge retention.

\begin{figure}
    \centering
    \includegraphics[width=1\linewidth]{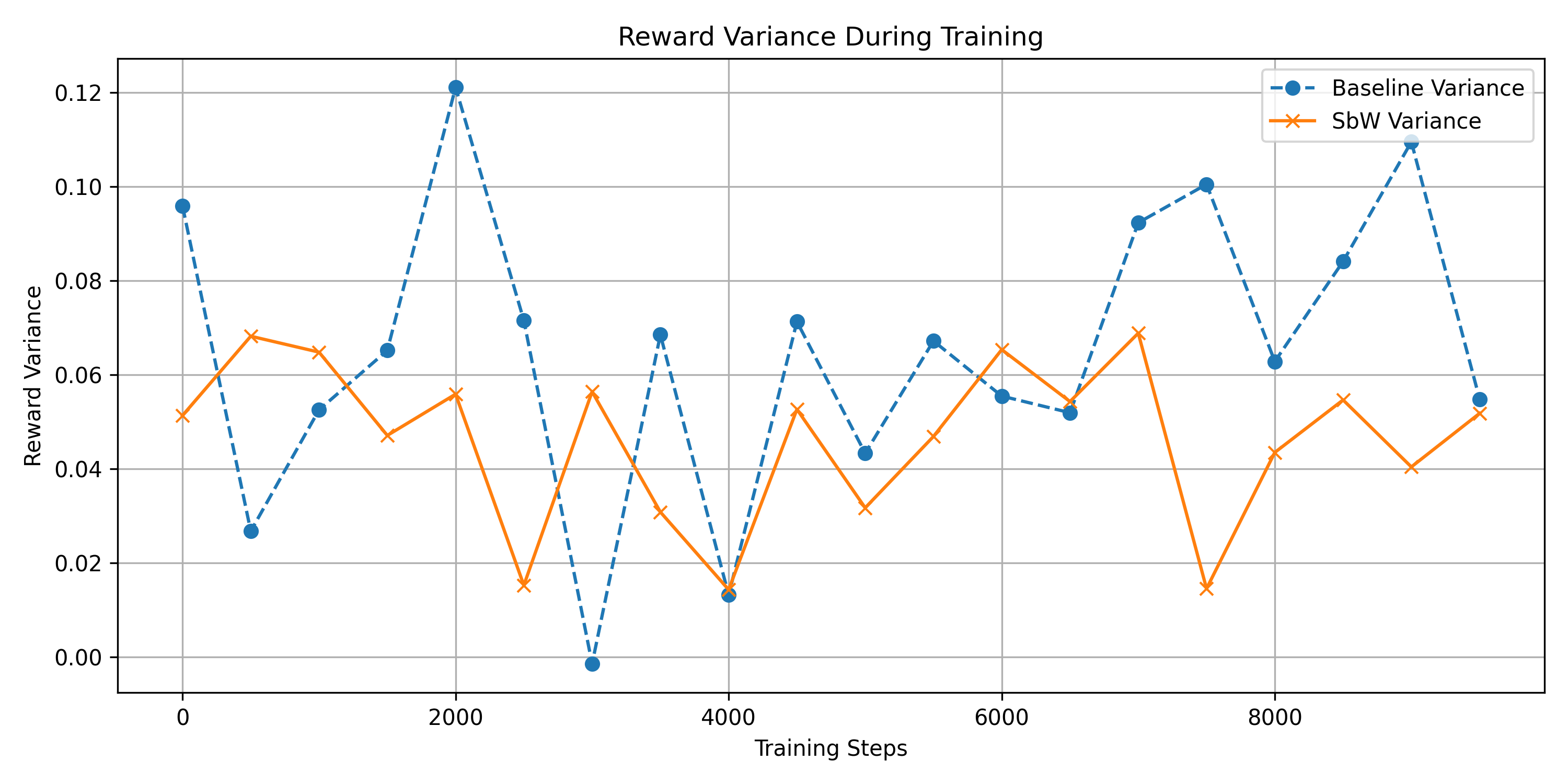}
    \caption{Comparison of training dynamics between the baseline and the proposed SbW strategy.}
    \label{fig:reward_variance}
\end{figure}

We further assess training dynamics by plotting the reward variance over training epochs in Figure~\ref{fig:reward_variance}. The model trained with SbW exhibits more stable convergence and lower reward variance, indicating that SbW acts as a regularizer that smooths the optimization landscape. These observations collectively demonstrate that SbW improves robustness and enhances reward-driven learning.

\subsection{Generalization}
\begin{table}[t]
\centering
\caption{Performance Comparison of SFT, DPO, and RL Paradigms on Verilog Code Generation Benchmarks}
\label{tab:training-paradigms}
\resizebox{\linewidth}{!}{
\begin{tabular}{lccc}
\toprule
\textbf{Model} & \textbf{VerilogEval} & \textbf{VerilogEval} & \textbf{RTLLM v1.1} \\
& \textbf{pass@1} & \textbf{pass@5} & \textbf{pass@5} \\
\midrule
CodeQwen1.5 & 22.5 & 26.1 & 24.1 \\
DeepSeek-Coder & 30.2 & 33.9 & 28.5 \\
CodeQwen2.5 & 31.0 & 43.0 & 29.3 \\
\midrule
SFT + CodeQwen1.5 & 38.4 & 50.5 & 46.2 \\
SFT + DeepSeek-Coder & 40.7 & 52.4 & 48.1 \\
SFT + CodeQwen2.5 & 42.1 & 56.4 & 53.3 \\
\midrule
DPO + CodeQwen1.5 & 40.3 & 51.3 & 58.4 \\
DPO + DeepSeek-Coder & 39.1 & 55.3 & 53.3 \\
DPO + CodeQwen2.5 & 47.6 & 60.2 & 56.7 \\
\midrule
VeriRL-DeepSeek-Coder (ours) & 66.7 & 71.6 & 62.4 \\
VeriRL-CodeQwen2.5 (ours) & \textbf{69.3} & 78.1 & 66.0 \\
Haven(ours) & 68.2 & \textbf{79.0} & \textbf{70.7} \\
\bottomrule
\end{tabular}
}
\end{table}
We compare our method against standard supervised fine-tuning (SFT), direct preference optimization (DPO), and open-source Verilog generation models to evaluate overall performance improvements. As shown in Table~\ref{tab:training-paradigms}, both SFT and DPO bring significant gains over pretrained models, yet they still struggle to fully exploit compiler-level and test-level feedback.

Our method—combining Trace-back based Rescore (TbR), Sample-balanced Weighting (SbW), and reinforce++ fine-tuning—consistently achieves superior results across backbones. For instance, with the CodeQwen2.5 backbone, our approach achieves 69.3\% pass@1 and 78.1\% pass@5 on VerilogEval, outperforming both SFT and DPO baselines. On the RTLLM v1.1 benchmark, we also observe strong generalization, with our method reaching 66.0\% pass@5.

In addition, we integrate our framework with Haven, an open-source Verilog code generation model, to test compatibility and effectiveness. Despite no architectural changes to Haven itself, applying our method boosts its pass@5 to 79.0\% on VerilogEval and 70.7\% on RTLLM, demonstrating the transferability and robustness of our reinforcement learning pipeline. These results underscore the broad applicability and general effectiveness of our approach.

\section{Conclusion}

We present \sysname, a reinforcement learning framework tailored for Verilog code generation. By addressing key challenges such as sparse rewards and training instability, our method combines a curated Veribench-53K dataset, a reasoning-aware Trace-back based Rescore mechanism, and a Sample-balanced Weighting strategy for stable fine-tuning. Experimental results demonstrate that \sysname outperforms existing SFT-based methods in both correctness and robustness, marking a significant step toward executable and semantically aligned HDL code generation.

\newpage
\bibliographystyle{IEEEtran}

\newpage
\appendix
\section{Appendix}

\definecolor{darkgreen}{rgb}{0,0.5,0}
\definecolor{keypink}{RGB}{204, 84, 205}
\definecolor{keybackblue}{RGB}{225, 232, 255}
\definecolor{keybackgreen}{RGB}{227, 234, 209}
\definecolor{keybackorange}{HTML}{FFD6A5}
\definecolor{keyblue}{HTML}{005FD7}
\definecolor{keygreen}{HTML}{005F00}
\definecolor{namepurple}{HTML}{5F00D7}
\definecolor{numberorange}{HTML}{D75F00}
\definecolor{llmblue}{RGB}{149, 206, 255}
\definecolor{vcspurple}{RGB}{207, 88, 238}
\definecolor{basefg}{RGB}{46,50,57}
\definecolor{keyblue}{RGB}{64,120,242}
\definecolor{keyred}{RGB}{212,80,68}
\definecolor{numberorange}{RGB}{202,119,38}
\definecolor{namepurple}{RGB}{155,89,182}
\definecolor{darkgreen}{RGB}{0,128,0}
\definecolor{gray}{gray}{0.5}

\lstdefinestyle{prompt}{
    language=,  
    basicstyle=\fontsize{6}{6}\ttfamily\color{black},  
    showstringspaces=false,
    breaklines=true,
    breakatwhitespace=false,
    emphstyle={[1]\color{keyblue}\bfseries},  
    emphstyle={[2]\color{keyred}\bfseries},            
    emph={[1]{PROBLEM, PREVSOLUTION, FEEDBACK, REASON, SOLUTION}},
    emph={[2]{Problem, Description, Previous, Attempt, Compiler, Testbench, Feedback}},
}

\lstdefinestyle{plain}{
    basicstyle=\fontsize{6}{6}\ttfamily,
    columns=fullflexible,
    keywordstyle=\color{blue},
    commentstyle=\color{gray},
    stringstyle=\color{green},
    showstringspaces=false,
    breaklines=true,
    breakatwhitespace=false,
    breakindent=0pt,
    escapeinside={(*@}{@*)},
    aboveskip=0pt,
    belowskip=0pt,
    xleftmargin=0pt,
}
\lstdefinestyle{verilog}{
    language=Verilog,
    basicstyle=\fontsize{6}{6}\ttfamily,
    keywordstyle=\color{keyblue},
    keywordstyle=[2]\color{black},
    commentstyle=\color{gray},
    stringstyle=\color{darkgreen},
    showstringspaces=false,
    breakatwhitespace=false,
    breaklines=true,
    breakindent=0pt,
    belowskip=0pt,
    escapeinside={(*@}{@*)},
    deletekeywords={reg, wire, always, begin, end},
    morekeywords=[2]{reg, wire},
    otherkeywords={always, begin},
    morekeywords=[3]{always, begin, end},
    keywordstyle=[3]\color{keypink},
    emphstyle={[1]\color{namepurple}},
    emph={[2]ns, ps},
    literate=
        {\'b}{{{\color{numberorange}'b}}}2
        {\'h}{{{\color{numberorange}'h}}}2
        {\'d}{{{\color{numberorange}'d}}}2
        {0}{{{\color{numberorange}0}}}1
        {1}{{{\color{numberorange}1}}}1
        {2}{{{\color{numberorange}2}}}1
        {3}{{{\color{numberorange}3}}}1
        {4}{{{\color{numberorange}4}}}1
        {5}{{{\color{numberorange}5}}}1
        {6}{{{\color{numberorange}6}}}1
        {7}{{{\color{numberorange}7}}}1
        {8}{{{\color{numberorange}8}}}1
        {9}{{{\color{numberorange}9}}}1
}

\begin{tcolorbox}[title=Generate Testbench Prompt, left=0mm, right=0mm, top=0mm, bottom=0mm,  colback=white, label=app:testbench-prompt,fonttitle=\footnotesize]
\begin{lstlisting}[style=plain]
You are the latest and most advanced bot designed to transform verilog code snippets and description into HDLBits-style questions. You will be provided with a Verilog-related function description along with a reference program that answers the question. Please perform the following tasks:

1. Come up with a HDLBits style question which consists of a well-defined problem. The generated question should meet the following criteria:
   a. The problem is clear and understandable, and contains the module head as well as the input and output
   b. The question should be solvable by implementing only one Verilog module. Avoid requiring multiple modules or complex pipelines.
   c. The problem should not require access to any external resources or databases.
   d. Feel free to use part of the original description if necessary, but ensure it is relevant for Verilog implementation.
   e. The question should be enclosed within <question></question> tags.
2. Based on the modified question generated in part 1, you need to generate a testbench for it, which needs to contain approximately 5 test cases, unless all possibilities have been covered by the cases. These test cases should cover both typical and boundary cases. Guidelines for writing the testbench and formatting:
   a. The **module under test** should be instantiated using the exact module name from the problem description (do not abbreviate or modify it).
   b.Ensure the **syntax and functionality** of the Verilog code are correct.
   c.**Enclose the entire testbench** within <testbench></testbench> tags.
   d.The **testbench should be self-contained**, meaning it should generate all required signals (clock, reset, inputs) internally without needing anything external
   e. Show the results of the execution of each test case (pass or failed) and do not terminate the program in the middle.
Below shows an example:
Description and verilog code:
\end{lstlisting}
\vspace{-2mm}
\begin{lstlisting}[style=verilog]

The Verilog code defines a module named `VCC` that outputs a constant high logic level (1). The output `V` is always set to 1.

module VCC (output V);
 assign V = 1'b1;
endmodule

Output:
<question>
Build a circuit that always outputs a HIGH.
module top_module (
  output V
);
</question>
<testbench>
module tb_module;
 wire V;

  // Instantiate the VCC module under test
  VCC uut (
    .V(V)
  );

  // Test case 1: Initialize and check that V is 1 at the start
  initial begin
    #1; // Wait for a moment to check the output
    if (V !== 1'b1) begin
      $display("Test Case 1 Failed: Expected V = 1, got V = %b", V);
    end else begin
      $display("Test Case 1 Passed: V = %b", V);
    end

''''''
// Test case 5: Check output at time = 100

    #100;
    if (V !== 1'b1) begin
      $display("Test Case 10  Failed: Expected V = 1, got V = %b", V);
    end else begin
      $display("Test Case 10 Passed: V = %b", V);
    end
 

    #150; // Wait for all test cases to finish
    $display("All test cases executed");
    $finish;
  end
endmodule
</testbench>
Now, generate the corresponding Verilog testbench.
description and verilog code:
```
{problem}
```
Output:
\end{lstlisting}
\end{tcolorbox}

\begin{tcolorbox}[title=TbR template, left=0mm, right=0mm, top=0mm, bottom=0mm,  colback=white, label=app:tbr-template, fonttitle=\footnotesize]
\begin{lstlisting}[style=plain]
You are an expert in digital logic design and Verilog programming.

You are given a Verilog problem description, a previous implementation attempt, and compiler/testbench feedback. Your task is to revise the Verilog code to fix any issues and improve the pass rate.

Carefully analyze the problem and identify why the previous solution failed. Then, generate a corrected Verilog implementation. Provide a clear explanation of what you changed and why.

Use the following format:
\end{lstlisting}
\vspace{-2mm}
\begin{lstlisting}[style=prompt]

<REASON>
[Explain the issues in the previous attempt based on the feedback and how you fixed them in the new implementation.]
</REASON>

<SOLUTION>
[Revised Verilog implementation]
</SOLUTION>

## Problem Description:
<PROBLEM>
[Insert the Verilog problem description here]
</PROBLEM>

## Previous Attempt:
<PREV_SOLUTION>
[Insert the previous Verilog implementation here]
</PREV_SOLUTION>

## Compiler/Testbench Feedback:
<FEEDBACK>
[Insert the compiler error messages and/or testbench pass rate here]
</FEEDBACK>

\end{lstlisting}
\end{tcolorbox}

\begin{tcolorbox}[title=TbR-Based Code Correction, left=1mm, right=1mm, top=0.5mm, bottom=0.5mm,  colback=white, fonttitle=\footnotesize]
\phantomsection
\label{case:sync_counter}
\vspace{-2mm}
\begin{lstlisting}[style=verilog]
## Problem Description:
<PROBLEM>
Design a 4-bit synchronous counter with active-high reset and enable.
</PROBLEM>

## Previous Attempt:
<PREV_SOLUTION>
module counter (
    input clk,
    input rst,
    input en,
    output reg [3:0] count
);
    always @(posedge clk or posedge rst) begin
        if (rst)
            count <= 4'b0000;
        else if (en)
            count <= count + 1;
    end
endmodule
</PREV_SOLUTION>

## Compiler/Testbench Feedback:
<FEEDBACK>
Compile successful. Pass rate: 3/10.
Error: Counter does not reset correctly when 'rst' is asserted at runtime.
</FEEDBACK>


<REASON>
Although the Verilog code compiles successfully, the reset behavior is not implemented correctly. Since this is a **synchronous counter**, reset should be checked only on the rising edge of the clock. However, the original implementation incorrectly uses `posedge rst` in the sensitivity list, making the reset **asynchronous**. This causes mismatch with the intended synchronous behavior and results in failed test cases. 

To fix this, we remove `posedge rst` from the sensitivity list and check the `rst` signal inside the `posedge clk` block.
</REASON>

<SOLUTION>
module counter (
    input clk,
    input rst,
    input en,
    output reg [3:0] count
);
    always @(posedge clk) begin  // <-- fixed: removed posedge rst
        if (rst)                 // synchronous reset
            count <= 4'b0000;
        else if (en)
            count <= count + 1;
    end
endmodule
</SOLUTION>

\end{lstlisting}
\end{tcolorbox}


\end{document}